\pdfoutput=1

\documentclass[11pt]{article}

\usepackage{acl}

\usepackage{times}
\usepackage{latexsym}
\usepackage{booktabs} 
\usepackage{graphicx}
\usepackage{placeins}
\usepackage{float}
\usepackage{subfigure}
\usepackage{amssymb}
\usepackage{amsmath}
\usepackage{multirow}
 \usepackage{longtable}
 
\usepackage[T1]{fontenc}

\usepackage[utf8]{inputenc}

\usepackage{microtype}

\title{Dependency-based Mixture Language Models}

\author{Zhixian Yang \and Xiaojun Wan \\
  Wangxuan Institute of Computer Technology, Peking University \\
  Center for Data Science, Peking University \\
  The MOE Key Laboratory of Computational Linguistics, Peking University \\
  \texttt{yangzhixian@stu.pku.edu.cn} \\
  \texttt{wanxiaojun@pku.edu.cn} \\}

\begin{document}
\maketitle
\begin{abstract}
Various models have been proposed to incorporate knowledge of syntactic structures into neural language models. However, previous works have relied heavily on elaborate components for a specific language model, usually recurrent neural network (RNN), which makes themselves unwieldy in practice to fit into other neural language models, such as Transformer and GPT-2. In this paper, we introduce the Dependency-based Mixture Language Models. In detail, we first train neural language models with a novel dependency modeling objective to learn the probability distribution of future dependent tokens given context. We then formulate the next-token probability by mixing the previous dependency modeling probability distributions with self-attention. Extensive experiments and human evaluations show that our method can be easily and effectively applied to different neural language models while improving neural text generation on various tasks.\footnote{Our code is available at \url{https://github.com/FadedCosine/Dependency-Guided-Neural-Text-Generation}}
\end{abstract}


\section{Introduction}

Syntactic structures serve as the principle of how words are correctly combined to form sentences. It is widely acknowledged that learning syntactic structures should improve neural text generation \cite{PRPN, PaLM, DU-SYD}. Even though current neural language models, such as Transformer \cite{transformer} and GPT-2 \cite{gpt-2} have achieved outstanding performance without explicitly modeling latent syntactic structures, these models still fail to learn the long-range syntactic dependencies \cite{LSTMFailStructure, Syntax-EnhancedPTM}.

To leverage explicit syntactic knowledge in natural language generation (NLG), many methods have been proposed \cite{seq2depNMT, PRPN, SIVAE, urnng, DU-SYD}. We conclude from previous works that knowledge of syntactic structures can bring four advantages to neural language models: 

(1) Syntactic structures can be modeled to obtain better representations of natural language sentences \cite{LearningHierarchicalStructures, DoLatentTree, TreeTransformer}.

(2) Jointly training syntactic structure parsing and language modeling can contribute to each other \cite{PRPN, rnng, urnng, DU-SYD, StructFormer}.

(3) Syntactic structures can be used to directly model the composition of language \cite{RNNforSentiment, SIELM} and help with the long-range dependency problem by providing shortcuts for gradient backpropagation \cite{HMRNN}.

(4) Integrating syntactic structures into a neural network can improve generalization via a better inductive bias \cite{onlstm, SIVAE}.

\begin{table*}[t]
\centering
\resizebox{\textwidth}{!}{
\begin{tabular}{cccc}
\toprule
Models              & External Parameters? & External Networks? & Architecture Agnostic? \\ \hline
RNNG  \cite{rnng}             & Yes                & Yes                   & No                     \\
PRPN    \cite{PRPN}           & Yes                & Yes                   & No                     \\
URNNG   \cite{urnng}            & Yes                & Yes                   & No                     \\
ON-LSTM  \cite{onlstm}          & Yes               &   No                 & No                     \\ \hline
DMLM (Ours) & No or Negligible                & No                    & Yes                      \\ 
\bottomrule
\end{tabular}}
\caption{\label{tab:model-comparison}
The difference between our DMLM and previous neural language models that incorporate knowledge of syntactic structures. Previous models often require external networks and external Parameters. For example, PRPN consists of three neural networks: Parsing Network, Reading Network and Predict Network. ON-LSTM is built upon a single LSTM, but it requires two additional gates in the LSTM cells, which leads to external parameters. All these previous models can only be built upon RNN architecture. However, as an architecture-agnostic method, DMLM needs no external parameters or networks when built upon Transformer, while it only needs negligible external parameters when built upon RNN.
}
\end{table*}

Despite these advantages, it is not trivial to incorporate knowledge of syntactic structures into neural language models effectively and efficiently. Several practical problems arise:

(1) Previous works \cite{HMRNN, PRPN, rnng, urnng, onlstm} have relied heavily on elaborate components for a specific language model, usually recurrent neural network (RNN) \cite{rnn}. These methods are difficult to be adapted to other neural language models, such as Transformer and GPT-2.

(2) If jointly modeling language modeling and syntactic structure parsing, it will require much more time/memory during training or inference.

To address these problems while keeping the advantages, we explore incorporating knowledge of syntactic structures in a different manner. In this work, we propose a novel dependency modeling objective to train neural language models to directly predict the current token's \textit{future dependent tokens} given the history. We define the \textit{future dependent tokens} of a specific token in a sentence as its children and parent in the dependency parse tree that will appear in the rest of the sentence. Further, we propose Dependency-based Mixture Language Models (DMLM) that, at each timestep, mixes the previous dependency modeling probability distributions with self-attention to get the next-token probability. As shown in Table~\ref{tab:model-comparison}, the proposed method can be adapted to any neural language model without adding external networks or parameters.

\begin{figure}[t]
	\begin{center}
		\centerline{\includegraphics[width=\columnwidth]{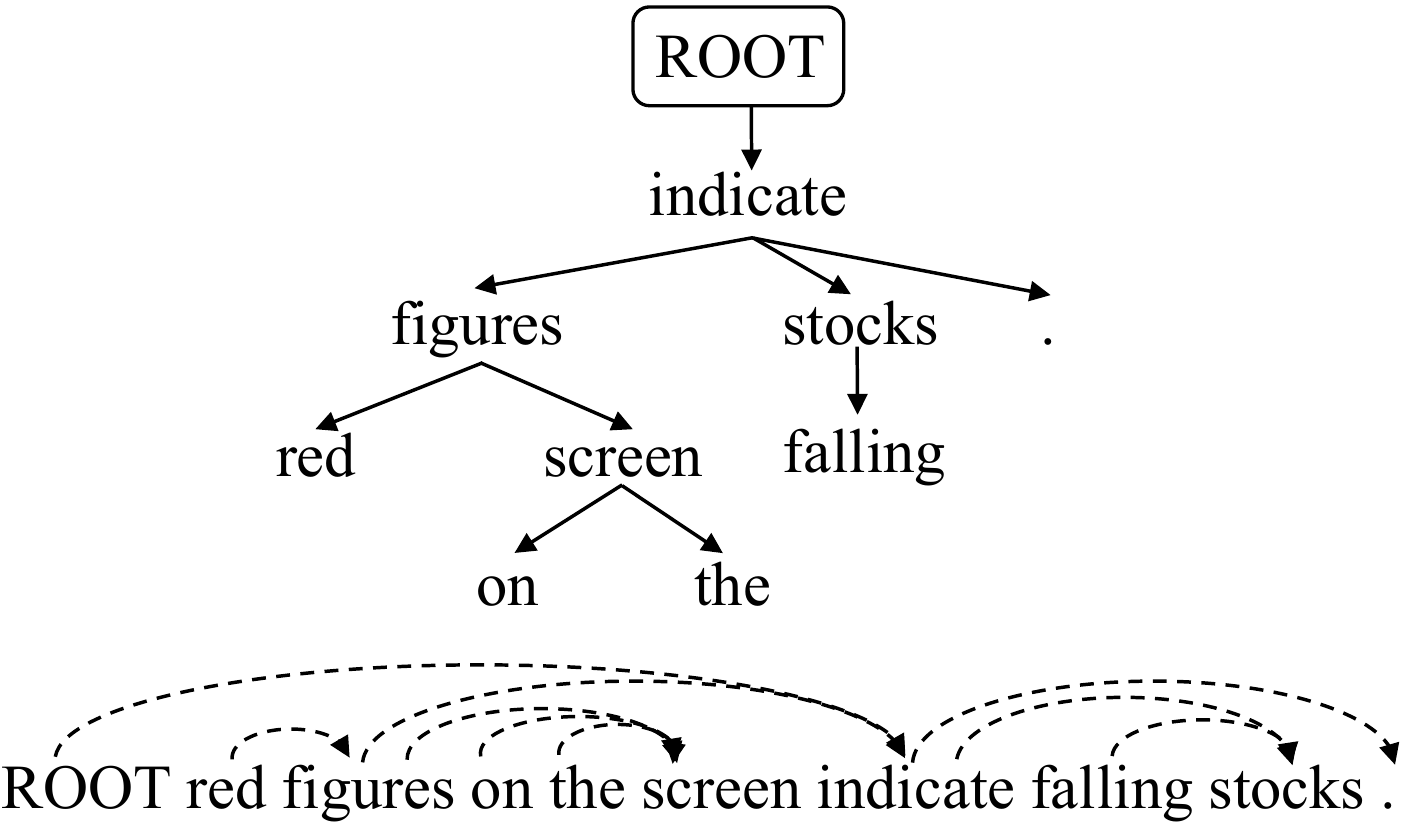}}
		\caption{Example of dependency parse tree}
		\label{fig-dep-modeling}
	\end{center}
	\vskip -0.1in
\end{figure}

\begin{figure*}[t]
	\begin{center}
		\centerline{\includegraphics[width=\textwidth]{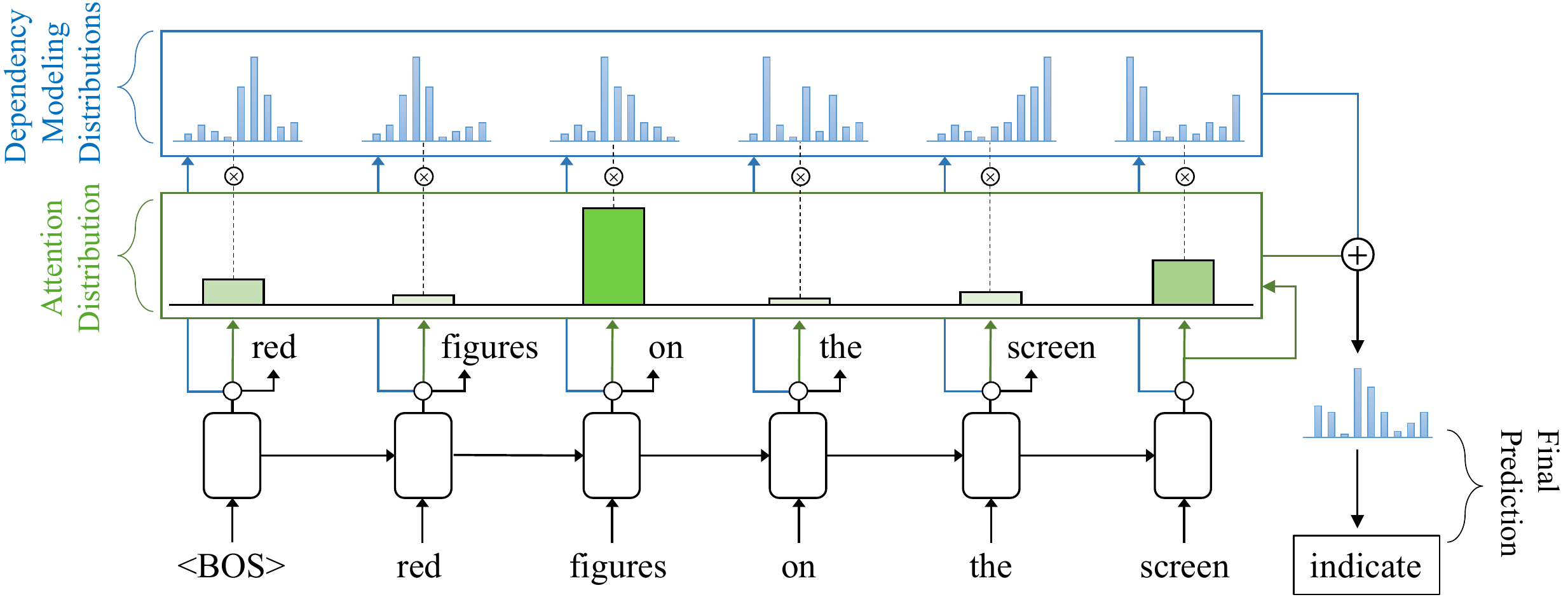}}
		\caption{Illustration of DMLM. For each timestep, the language model outputs a dependency modeling distribution, while the self-attention produces a dependency attention distribution over the context. And then, the next-token probability is the sum of the context's dependency modeling probability distributions weighed by the dependency attention scores. Best viewed in color.}
		\label{fig-DMLM}
	\end{center}
	\vskip -0.2in
\end{figure*}

Our core idea can be illustrated in Figure~\ref{fig-dep-modeling} and Figure~\ref{fig-DMLM}: when predicting the next-token "indicate" after reading "red figures on the screen", common language models are easy to predict an incorrect word, such as "indicates", since the prediction of these models relies heavily on the recent word, "screen" in this case. However, our propose DMLM will directly look back into the long-range context, and select the next-token from all the future dependent tokens predicted by previous tokens. According to the underlying dependency structure, DMLM pays different weights to different tokens' future dependent tokens. Thus, the model is more likely to predict "indicate" since DMLM tends to think of the next-token as a future dependent token of "figures" rather than "screen".


We conduct experiments with different neural language models including LSTM \cite{LSTM}, Transformer \cite{transformer}, and GPT-2 \cite{gpt-2} across different tasks in conditional text generation, unconditional text generation, and language modeling. Through extensive experiments we demonstrate that DMLM consistently improves the generation quality according to both human evaluations and automatic metrics. Compared to other neural language models that incorporate syntactic knowledge, DMLM is architecturally simpler and easier to fit into any neural language model, while possessing wide applicability to different text generation tasks.

\section{Methodology}

Our goal is to propose a simple yet effective method that can improve neural text generation by learning from the underlying syntactic structure, and can fit into any auto-regressive generation model without using additional elaborate components. We first introduce a novel dependency modeling objective to force the model to directly predict the future dependent tokens of the current token. Based on the dependency modeling, we then present the proposed DMLM.

\subsection{Dependency Modeling}
\label{dependency_modeling}

It has been a challenge to equip neural language models with the capability of modeling long-range dependency in text \cite{transformer-xl}. In particular, previous works \cite{seq2depNMT} observe that vanilla RNN can hardly capture many subtle long-range token dependencies effectively. On the other hand, though self-attention mechanisms can build direct connections between long-distance token pairs, it is still elusive for Transformer to be aware of syntactic dependency structures while also obtaining strong language modeling performance \cite{SOM}.

The current neural language models are mostly trained purely using the language modeling objective with Maximum Likelihood Estimation (MLE). With the auto-regressive factorization, language modeling can be reduced to modeling the conditional distribution of the next-token $x_t$ given the context $\mathbf{x}_{<t}=\{x_1, \dots , x_{t-2}, x_{t-1} \}$. However, in order to make neural language models aware of long-range dependency and syntactic structures, we propose the dependency modeling objective to train models to learn the probability distribution of the future dependent tokens directly. Following \citet{YouOnlyNeedTraverseTrees}, we define the \textit{future dependent tokens} of a specific token in a sentence as its children and parent in the dependency parse tree that will appear in the rest of the sentence. Taking Figure~\ref{fig-dep-modeling} as an example, the future dependent tokens of "figures" are "screen" and "indicate", since "red" does not appear after "figures" in this sentence.

Specifically, given a token sequence $\mathbf{x} = \{ x_1, \dots, x_{T-1}, x_T \} $ where $T \in \mathbb{N}$ denotes the sequence length, we first use dependency parser to generate a dependency tree. Then, we derive the future dependent tokens set $Z_t$ for each token $x_{t-1}$, where $Z_t = \{ x_i \mid i \ge t, x_i \text{ is the child or parent of } x_{t-1} \}$. We train a language model $\theta$ to maximize the log-likelihood sum of tokens in $Z_t$. This equals to minimize:
\begin{equation}
	\label{eq-dependency-modeling}
	\mathcal{L}_{\mathrm{DM}}\left(\theta\right)=-\sum_{t=1}^{T}\sum_{z_t \in Z_t} \log p^{\text{dep}}_{\theta}\left(z_t \mid \mathbf{x}_{<t}\right),
\end{equation}
which is the dependency modeling objective.

\subsection{Dependency-based Mixture Language Models}

To give a categorical probability distribution over the next-token, a standard approach for the current neural language models is to encode the context into a fixed-size vector followed by an output embedding layer and a softmax function. 

In our case, given the context $\mathbf{x}_{<t}$, we first train the language model to directly learn the probability distribution of $x_{t-1}$'s future dependent tokens $p^{\text{dep}}_{\theta}\left(w \mid \mathbf{x}_{<t}\right)$ by dependency modeling (Section~\ref{dependency_modeling}). We then propose DMLM (depicted in Figure~\ref{fig-DMLM}) that mixes dependency modeling probability distributions $P^{\text{dep}} = \{p^{\text{dep}}_{\theta}\left(w \mid \mathbf{x}_{<1}\right), \dots , p^{\text{dep}}_{\theta}\left(w \mid \mathbf{x}_{<t-1}\right),$ $p^{\text{dep}}_{\theta}\left(w \mid \mathbf{x}_{<t}\right)\}$. All the probability distributions in $P^{\text{dep}}$ are weighed by self-attention, and summed to obtain the final next-token probability distribution.


We can easily implement a self-attention in both Transformer-based and RNN-based language models. For example, in Transformer and GPT-2, the penultimate layer seems to naturally learn alignments \cite{AlignTransformer}, so we use its average attention weights over all the attentions heads as the dependency attention distribution. In RNN-based models, inspired by \citet{PointerRNN} and \citet{transformer}, at each timestep, we linearly project the current hidden state $h_t \in \mathbb{R}^H$ to a query vector $q_t = W^{Q} h_t$ and a key vector $k_t = W^{K} h_t$, where $W^{Q} \in \mathbb{R}^{H \times H}$, $W^{K} \in \mathbb{R}^{H \times H}$, $q_t \in \mathbb{R}^H$, and $k_t \in \mathbb{R}^H$. To generate the dependency attention, we compute the match between the query $q_t$ and the context's keys $\{k_{1}, \dots, k_{t-1}, k_{t}\}$  by taking the inner product, followed by a softmax to obtain the dependency attention distribution:
\begin{equation}
\begin{aligned}
	\label{eq-dependency-attention}
	\mathbf{e^{\text{(t)}}} &= \{e^{(t)}_{1}, \dots, e^{(t)}_{t-1}, e^{(t)}_{t} \}, \\
	e^{(t)}_i & = q_t^T k_i, 1 \le i \le t, \\
	\mathbf{a^{(t)}} &= \text{softmax} (\frac{\mathbf{e^{\text{(t)}}}}{\sqrt{H}}), \\
	\mathbf{a^{(t)}} & = \{a^{(t)}_{1}, \dots, a^{(t)}_{t-1}, a^{(t)}_{t} \},
\end{aligned}    
\end{equation}
where $\mathbf{e^{\text{(t)}}} \in \mathbb{R}^t$, and $\mathbf{a^{(t)}} \in \mathbb{R}^t$. We scale the dot products by $\frac{1}{\sqrt{H}}$ following \citet{transformer}. 

 
The dependency attention distribution reveals which token in the context may have a strong dependency relation with the token to be predicted. Thus, the neural language model should pay more attention to previous tokens with high dependency attention scores, i.e., the next-token is more likely to be the future dependent token of those tokens in the context. Formally, the next-token probability is the sum of the context's dependency modeling probability distributions weighed by the dependency attention scores:
\begin{equation}
	\label{eq-dependency-weighed-distribution}
	p_{\theta}\left(w \mid \mathbf{x}_{<t}\right)=\sum_{\tau=1}^{t}a^{(t)}_\tau p^{\text{dep}}_{\theta}\left(w \mid \mathbf{x}_{<\tau}\right).
\end{equation}
where $p^{\text{dep}}_{\theta}\left(w \mid \mathbf{x}_{<\tau}\right)$ is the probability distribution of $x_{\tau-1}$'s future dependent tokens, since till now the neural language model is only trained by dependency modeling. Then, we further finetune the neural language model using MLE, but with respect to our modified probability distribution given in Equation~\ref{eq-dependency-weighed-distribution}:
\begin{equation}
	\label{eq-MLE-finetune}
	\mathcal{L}_{\mathrm{LM}}\left(\theta\right)=-\sum_{t=1}^{T} \log p_{\theta}\left(x_t \mid \mathbf{x}_{<t}\right).
\end{equation}


For each timestep during inference, DMLM outputs a dependency modeling distribution, and we store it in a list. To predict the next-token, DMLM applies self-attention in Equation~\ref{eq-dependency-attention} to produce a dependency attention distribution over the context, and then the next-token probability can be calculated by Equation~\ref{eq-dependency-weighed-distribution}, where the list preserves all the $p^{\text{dep}}_{\theta}\left(w \mid \mathbf{x}_{<\tau}\right), 1\le \tau \le t$. 

\begin{table*}[t]
\centering
\begin{tabular}{l|llll|llll}
\toprule
Models & UNION $\uparrow$& BERTScore $\uparrow$  & B-1  $\uparrow$ & B-2  $\uparrow$& D2   $\uparrow$ & D3  $\uparrow$  & SB-2 $\downarrow$ & SB-3 $\downarrow$ \\ \hline
PRPN  & 83.37 & 29.11 & 21.45 & 6.84 & 13.22 & 33.50 & 95.17 & 86.76  \\
ON-LSTM  & 82.18 & 29.41 & 22.16 & 7.33 & 13.93 & 35.71 & 94.98 & 85.80  \\
AWD-LSTM  & 82.98 & 29.57 & 22.23 & 7.31 & 14.07 & 35.71 & 94.92 &  85.88 \\
DM-LSTM        & \textbf{83.97}$^\star$ & \textbf{29.93} & \textbf{22.54}$^\star$ & \textbf{7.63}$^\star$ & \textbf{14.92} & \textbf{37.44} & \textbf{94.47}$^\star$ & \textbf{84.77}$^\star$ \\ \hline
Transformer    & 81.39 & 27.64 & 21.28 & 7.01 & 17.48 & \textbf{42.30}  & 93.18 & 81.52 \\
DM-Transformer & \textbf{84.07}$^\star$ & \textbf{28.20}$^\star$ & \textbf{21.49} & \textbf{7.29}$^\star$ & \textbf{17.79} & 42.08 & \textbf{92.86}$^\star$ & \textbf{81.36}$^\star$ \\ \hline
GPT-2 & 84.41 & 29.02 & 21.79 & 7.45 & 17.09 & 40.74 & 93.51 & 82.55 \\
DM-GPT-2       & \textbf{85.31}$^\star$ & \textbf{30.18}$^\star$ & \textbf{22.81}$^\star$ & \textbf{8.02}$^\star$ & \textbf{17.98} & \textbf{43.29} & \textbf{93.18} & \textbf{81.41}$^\star$ \\ \bottomrule
\end{tabular}
\caption{\label{tab:ctg-auto-result} Automatic evaluation results for the conditional text generation task on Rocstories dataset. $^\star$ denotes that DM-model significantly outperforms the second best model for $t$-test ($p$-value<0.05). }
\end{table*}

\begin{table*}[t]
\centering
\small
\begin{tabular}{l|llll|llll}
\toprule
\multirow{2}{*}{Models}        & \multicolumn{4}{c|}{Grammaticality}   & \multicolumn{4}{c}{Logicality}       \\
         & Win(\%) & Lose(\%) & Tie(\%) & $\kappa$   & Win(\%) & Lose(\%) & Tie(\%) & $\kappa$      \\ \hline
DM-LSTM vs. PRPN & 36.2$^\star$ & 14.5 & 49.3 & 0.225 &  56.5$^\star$ & 17.5 & 26.0 & 0.306\\   
DM-LSTM vs. ON-LSTM               & 12.8$^\star$  & 6.4      & 80.8    & 0.238 & 48.4$^\star$  & 24.4     & 27.2    & 0.409 \\
DM-LSTM vs. AWD-LSTM  & 28.0$^\star$ & 14.5 & 57.5 & 0.224 & 43.0$^\star$ & 34.5 & 22.5 & 0.214 \\
DM-Transformer vs. Transformer & 18.2$^\star$   & 5.2      & 76.6    & 0.358 & 50.6$^\star$  & 18.6     & 30.8    & 0.342 \\
DM-GPT-2 vs. GPT-2             & 20.4$^\star$   & 5.0        & 74.6    & 0.374 & 50.6$^\star$  & 18.8     & 30.6    & 0.224 \\
\bottomrule
\end{tabular}
\caption{\label{tab:ctg-HE} Human evaluation results for the conditional text generation task on Rocstories dataset. $\kappa$ denotes the inter-annotator agreement Krippendorff's alpha~\cite{Krippendorff} score. $^\star$ means statistical significance for Wilcoxon signed-rank test ($p$-value<0.01). Note that, it is relatively easy for both models to generate a single sentence that is grammatically correct, so the rate of "tie" in Grammaticality is relatively high. }
\vskip -0.1in
\end{table*}

\section{Experiments}

Despite previous works mainly focusing on language modeling, it has always been a thorny issue whether better language models lead to better performance in downstream tasks. Therefore, we showcase the performance of our proposed DMLM in three different tasks: conditional text generation (Section~\ref{CTG}), unconditional text generation (Section~\ref{UTG}), and language modeling (Section~\ref{LM}).

To verify the effectiveness and architecturally generalizability of our method, we conduct the generation tasks with three dominant neural language models, including LSTM, Transformer and GPT-2. We prefix the base model name with "\textbf{DM-}" to denote the corresponding Dependency-based Mixture language model. Specifically, we adopt AWD-LSTM~\cite{AWDLSTM} as our base LSTM, and further compare our DM-LSTM with \textbf{PRPN}~\cite{PRPN} and \textbf{ON-LSTM}~\cite{onlstm} which also incorporate knowledge of syntactic structures, and are built on LSTM. In the same task, we use exactly the same hyper-parameters and setups for the pairs of base models and corresponding DM-models. Other details of the experimental setup for each task can be seen in Appendix~\ref{DetailSetup}. 

For all the tasks, we use a state-of-the-art parser, HPSG Parser\footnote{\url{https://github.com/DoodleJZ/HPSG-Neural-Parser}}~\cite{HPSGParser} to get the dependency parse tree for each sentence in the datasets. We discuss the impact of the dependency parser in Appendix~\ref{impact_parser}.

\subsection{Conditional Text Generation}
\label{CTG}

\textbf{Setup}\quad We take the story ending generation as the conditional text generation task, and evaluate our method on the ROCStories corpus \cite{rocstories}, which consists of 98,161 five-sentences. We follow the preprocessing\footnote{We use the preprocessed data in https://github.com/thu-coai/Stylized-Story-Generation-with-Style-Guided-Planning} of  \citet{StylizedStory} to randomly split ROCStories by 8:1:1 for training/validation/test, respectively, and delexicalize stories by masking all the male/female/unknown names with "[MALE]"/"[FEMALE]"/"[NEUTRAL]". We finally get a word-level  vocabulary with $31,216$ unique tokens. The conditional text generation task is to generate a reasonable ending given a four-sentence story context. For all models, we generate stories using nucleus sampling~\cite{topp} with $p = 0.5$. 

We measure the generated story endings by the following automatics metrics: (1) \textbf{UNION}~\cite{UNION}: It is a learnable unreferenced metric for evaluating the quality of generated stories; (2) \textbf{BERTScore}~\cite{BERTScore}: The metric measures the semantic consistency between the generated and the referenced ones by BERT~\cite{BERT}; (3) \textbf{BLEU (B-n)}~\cite{bleu}: BLEU evaluates $n$-gram overlap between the generated stories and the references; (4) \textbf{Distinct (D-n)}~\cite{distinct}: The proportions of distinct $n$-grams in the outputs to evaluate the diversity of generated results. Since Distinct score will become extremely low for small $n$, we calculate it with $n=2,3$; (5) \textbf{Self-BLEU (SB-n)}~\cite{self-bleu}: The metric is calculated by computing $n$-grams ($n=2,3$) BLEU score of each generated text with all other generated ones as references. Smaller Self-BLEU scores indicate better diversity.

\noindent{\textbf{Results}
} \quad The experimental results of baselines and corresponding DM-models  are shown in Table~\ref{tab:ctg-auto-result}. Note that we do not conduct significant tests on Distinct since it is a document-level metric. We can see that, all the DM-models significantly outperform baseline models on almost all the metrics. Furthermore, compared with PRPN and ON-LSTM, our DM-LSTM performs significantly better in all the metrics. This indicates that incorporating knowledge of syntactic structures in our proposed way can effectively contribute to both the quality and diversity of the story ending generation. Moreover, no matter what the base model is, our DM-model can substantially improves the conditional text generation. This demonstrates that our method can be effectively adapted to different neural language models, such as the large scale language model, GPT-2, while previous models like ON-LSTM can only be built on LSTM. 

\begin{table}[t]
\centering
\begin{tabular}{l|cc}
\toprule
Models         & LM score  $\downarrow$    & RLM score  $\downarrow$   \\ \hline
PRPN &  5.24 & 5.75 \\
ON-LSTM           & 5.20           & 5.59          \\
AWD-LSTM & 5.18 & 5.64 \\
DM-LSTM        & \textbf{5.14} & \textbf{5.52} \\ \hline
Transformer    & 5.00             & 5.59          \\
DM-Transformer & \textbf{4.97} & \textbf{5.49} \\ \hline
GPT-2          & 4.89          & 5.55          \\
DM-GPT-2       & \textbf{4.67} & \textbf{5.47} \\ \bottomrule
\end{tabular}
\caption{\label{tab:utg-lmrlm-result} Results of global metrics for the unconditional text generation task on EMNLP2017 WMT News. }
\vskip -0.1in
\end{table}

\begin{table*}[t]
	\centering
	\begin{tabular}{l|llllllll}
		\toprule
		\multirow{2}{*}{Models} & \multicolumn{8}{c}{Nucleus-$p$} \\ \cline{2-9} 
		& 0.3   & 0.4   & 0.5   & 0.6   & 0.7   & 0.8   & 0.9    & 1.0    \\ \hline
		PRPN    & 41.48  &  45.77  &  55.32  & 64.23   &  83.98  & 109.3   & 172.09   &  302.57  \\ 
		ON-LSTM     & 37.46 & 42.98 & \textbf{46.16} & 56.69 & 72.36 & 98.06 & 152.60  & 274.43 \\
		AWD-LSTM & 37.97  &  41.80  &  48.74  & 57.45   &  71.77  & \textbf{94.22}   &  146.40  &  289.13  \\ 
		DM-LSTM  & \textbf{36.11} & \textbf{39.53}$^\star$ & 47.67 & \textbf{55.30}  & \textbf{69.38} & 95.95 & \textbf{136.98}$^\star$ & \textbf{256.51}$^\star$ \\ \hline
		Transformer    & 45.37 & 46.36 & 50.90  & 60.27 & 70.74 & 91.65 & 125.46 & 222.27 \\
		DM-Transformer & \textbf{37.74}$^\star$ & \textbf{40.75}$^\star$ & \textbf{43.25}$^\star$ & \textbf{49.92}$^\star$ & \textbf{60.28}$^\star$ & \textbf{76.77}$^\star$ & \textbf{104.03}$^\star$ & \textbf{182.29}$^\star$ \\ \hline
		GPT-2    & 41.19 & 44.05 & 47.86 & 53.97 & 63.18 & 81.45 & 112.81 & 192.10 \\
		DM-GPT-2 & \textbf{36.41}$^\star$ & \textbf{40.99}$^\star$ & \textbf{41.75}$^\star$ & \textbf{46.18}$^\star$ & \textbf{55.36}$^\star$ & \textbf{67.97}$^\star$ & \textbf{92.22}$^\star$  & \textbf{152.98}$^\star$ \\ \bottomrule
	\end{tabular}
\caption{\label{tab:utg-ppl-result} GPT-2 Perplexity on $1,000$ random samples with various sampling hyper-parameters generated by models trained on EMNLP2017 WMT News dataset. Nucleus sampling is used here with various $p$. $^\star$ denotes that DM-model significantly outperforms the second best model for $t$-test ($p$-value<0.05). }
\vskip -0.1in
\end{table*}
\begin{table}[t]
\centering
\begin{tabular}{l|c}
\toprule
Models         & Human score $\uparrow$   \\ \hline
PRPN          & 0.380       \\
ON-LSTM           & 0.278       \\
AWD-LSTM           & 0.365       \\
DM-LSTM        & \textbf{0.444}      \\ \hline
Transformer    & 0.400       \\
DM-Transformer & \textbf{0.448}       \\ \hline
GPT-2          & 0.468       \\
DM-GPT-2       & \textbf{0.512}        \\ \hline
Real data      & 0.688     \\
\bottomrule 
\end{tabular}
\caption{\label{tab:utg-HE} Turing test results of the samples generated by models trained on EMNLP2017 WMT News dataset. To reach a good trade-off between quality and diversity, we adopt nucleus sampling with $p=0.7$ for all the models to generate samples. }
\vskip -0.1in
\end{table}

\noindent{\textbf{Human evaluation}
} \quad To further evaluate the fluency and logic of generated stories, following~\cite{TACLStoryGen}, we conduct pair-wise comparisons between DM-models and corresponding baselines. We randomly sample 100 story endings from each model. For each pair of stories (one by the DM-model and the other by the baseline, along with the beginning), five annotators are hired to give a preference (win, lose, or tie) from the following two aspects: (1) \textit{Grammaticality}: whether a story ending is natural and fluent; (2) \textit{Logicality}: whether a story is coherent to the given beginning and reasonable in terms of causal and temporal dependencies in the context. The detailed questionnaire and other details are shown in Appendix~\ref{HEQuestion}.

The average win/lose/tie rates of the human evaluation are shown in Table~\ref{tab:ctg-HE}. To measure the inter-annotator agreement, we calculate Krippendorff's alpha~\cite{Krippendorff} for each pair-wise comparison, and all the results are fair agreement ($0.2 \le \kappa \le 0.4$) or moderate agreement ($0.4 \le \kappa \le 0.6$). The results show that our DM-models significantly outperform baseline models in both the grammaticality and logicality.

\subsection{Unconditional Text Generation}
\label{UTG}

\textbf{Setup}\quad We perform experiments of unconditional text generation on EMNLP2017 WMT News dataset\footnote{\url{
http://statmt.org/wmt17/translation-task.htm}}. We use the preprocessed data of a recent work\footnote{\url{https://github.com/pclucas14/GansFallingShort/tree/master/real_data_experiments/data/news}}~\cite{GanFallingShort} that contains $5,268$ distinct words with maximum sentence length 51. The training/validation/test set consists of $268,586/10,000/10,000$ sentences. 

Following \citet{GanFallingShort}, we evaluate the models with the \textit{global metrics} \cite{lm_rlm}: (1) \textbf{Language Model score (LM score)}: We use the oracle Language Model to evaluate the negative log-likelihood of generated text as the metric to reflect quality; (2) \textbf{Reverse Language Model score (RLM score)} We train a new Language Model on the generated text, and then evaluate the negative log-likelihood of a held-out set of real text. This metric can measure text diversity since the generated text with better diversity would have a broader coverage over the real data space, and the new Language Model can be trained better, thus leading to lower RLM score. Both the LM score and RLM score are usually evaluated on the sentences generated by purely random sampling. Besides, to further measure the generation fluency, we directly use the public GPT-2 checkpoint of pretrained parameters without finetuning to calculate \textbf{GPT-2 Perplexity} of generated samples. 

\begin{table*}[t]
\centering
\begin{tabular}{l|lll}
\toprule
Models                & \#Params & Dev PPL & Test PPL \\ \hline
Pointer Sentinel-LSTM \cite{PointerRNN} & 21M      & 72.4    & 70.9     \\
RNNG    \cite{rnng}              & -        & -       & 88.7     \\
Variational RHN  \cite{RHN}     & 23M      & 67.9    & 65.4     \\
PRPN     \cite{PRPN}             & -        & -       & 62.0       \\
Fraternal dropout \cite{FraternalDropout}    & 24M      & 58.9    & 56.8     \\
URNNG     \cite{urnng}            & -        & -       & 85.9     \\
ON-LSTM    \cite{onlstm}           & 25M      & 58.3    & 56.2     \\ \hline
AWD-LSTM   \cite{AWDLSTM}           & 24M      & 60.0      & 57.3     \\
DM-LSTM (Ours)              & 24M      & 58.6    & 56.2   \\ \hline
AWD-LSTM-MoS\cite{MoS}               & 22M      & 56.5    & 54.4 \\ 
AWD-LSTM-DOC\cite{DOC}               & 23M      & 54.1    & 52.4 \\
\bottomrule
\end{tabular}
\caption{\label{tab:lm-ppl}  Various language models' perplexity evaluated on validation and test sets of Penn Treebank dataset. \citet{MoS} and \citet{DOC} focus on improving the softmax of LSTM LM, which are orthogonal to ours. }
\vskip -0.1in
\end{table*}

\noindent{\textbf{Results}} \quad Table~\ref{tab:utg-lmrlm-result} shows the results of global metrics obtained by various models. All the DM-models again outperform the baselines. The consistently lower LM scores indicate that the generated sentences of DM-models are of better quality, while the consistently lower RLM scores also demonstrate that DM-models can generate more diverse sentences meanwhile. 
	
In addition, each model is used to generate $1,000$ sentences with various sampling hyper-parameters, and GPT-2 Perplexity is further calculated. As shown in Table~\ref{tab:utg-ppl-result}, our proposed method can make neural language models perform significantly better in terms of generation fluency. In particular, Transformer-based models can gain more significant improvement from DMLM. We conjecture that this is because, in our implementation, we directly uses the penultimate multi-head attention layer of Transformer to obtain the dependency attention distribution of DMLM. Thus, it can easily inherit all the strengths of Transformer-based models.
	
\noindent{\textbf{Human evaluation}
} \quad Following previous work~\cite{SeqGAN, LeakyGAN}, we conduct a Turing test to further evaluate the generated text. In practice, we mix $100$ randomly sampled sentences from each model, and another $100$ sentences from the real test set. Five annotators are hired to judge whether each of the $900$ sentences is created by human or machines. Each sentence gets $+1$ score when it is regarded as a real one, and $0$ score otherwise. The detailed questionnaire and other details are shown in Appendix~\ref{HEQuestion}.

The average score for each model is shown in Table~\ref{tab:utg-HE}, from which we can see all the DM-models surpass the baselines. Both automatic evaluations and human evaluations indicate that DMLM can help neural language models generate more readable, fluent, and natural sentences.

\subsection{Language Modeling}
\label{LM}

\noindent{\textbf{Setup}
} \quad We evaluate the proposed method with the word-level language modeling task by measuring \textbf{Perplexity (PPL)} on the Penn Treebank (PTB) \cite{PTB1, PTB2} corpora. The PTB dataset has a vocabulary size of $10,000$ unique words, and the training/validation/test set consists of $42,068/3,370/3,761$ sentences. 

For this task, we mainly implement the DMLM on the RNN-based language model, i.e., AWD-LSTM~\cite{AWDLSTM}. For a fair comparison, our DM-LSTM uses exactly the same hyper-parameters and setups as AWD-LSTM. Since Transformer-based models' strong performance relies on training with large datasets, it will perform worse than random when trained on a small dataset~\cite{SOM}. We still report Transformer-based models' language modeling results on PTB in Appendix~\ref{Transformer-lm}.

\noindent{\textbf{Results}
} \quad We compare our method with its base model, AWD-LSTM, and we report the results along with other state-of-the-art models in Table~\ref{tab:lm-ppl}. Compared with the AWD-LSTM, our DM-LSTM reduces the perplexity by 1.4 on the validation set and 1.1 on the test set, indicating that incorporating knowledge of syntactic structures in our proposed manner can substantially improve language modeling. Compared with other models that also leverage syntactic knowledge, our DM-LSTM strongly outperforms RNNG, PRPN, and URNNG. Moreover, though DM-LSTM does not make any changes to the architecture of the AWD-LSTM language model, it still achieves a comparable perplexity with ON-LSTM. Note that, since our method is model-agnostic, it can be harmonically combined with other state-of-the-art models, such as MoS~\cite{MoS} and DOC~\cite{DOC}.

\section{Discussion}

\subsection{Visualization}

\begin{figure}[t]
	\begin{center}
		\centerline{\includegraphics[width=\columnwidth]{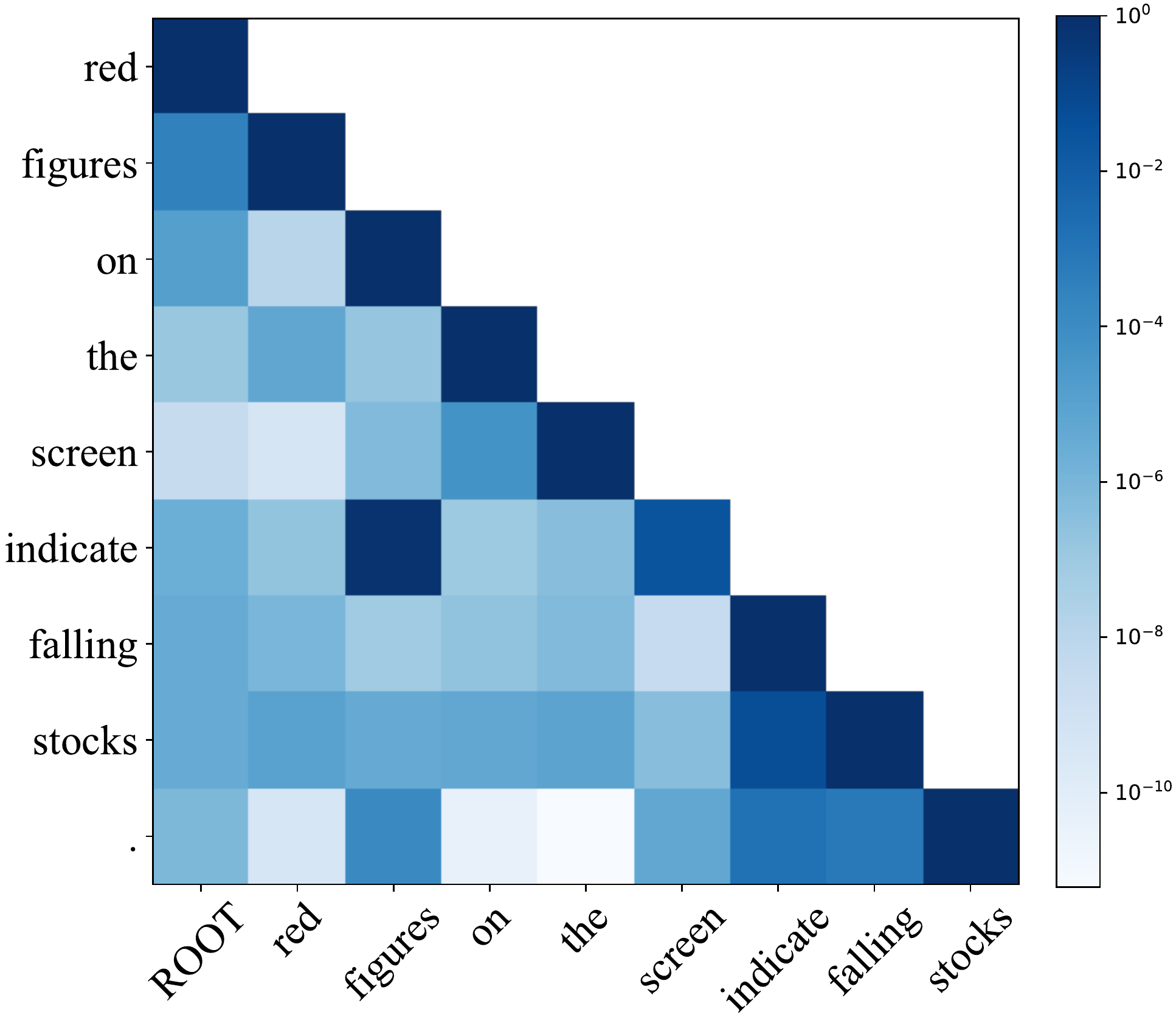}}
		\caption{Visualization of dependency attention distributions. We left-shift the sentence by one step in the y-axis to better display the attention between the predicted next-token and the context in each row. }
		\label{fig-AttentionMatrix}
	\end{center}
	\vskip -0.2in
\end{figure}

We show how our proposed method works by visualizing the dependency attention distributions. We use DM-Transformer to generate a sentence: "red figures on the screen indicate falling stocks." For each generation step, we record this step's dependency attention distribution. When we finally generate the whole sentence, we get $9$ distributions and plot Figure~\ref{fig-AttentionMatrix} from them.  Each row in Figure~\ref{fig-AttentionMatrix} shows the dependency attention distribution of the model when generating the corresponding Y-axis token. When predicting the token "indicate", DMLM pays great attention to "figures". This is because these two tokens have a direct dependency connection in the dependency parse tree, and our method successfully captures this relationship. In addition, DMLM also helps the model better organize dependency information when the next-tokens, such as "screen" and "stocks", have dependencies on more than one token in the context. 

\subsection{Case Study}
We perform case studies for a better understanding of the model performance. Table~\ref{Example-CTG} provides examples of conditional text generation produced by our DM-models and other baselines. Obviously, all the DM-models can generate more reasonable and coherent story endings. Additionally, some examples of unconditional text generation are shown in Table~\ref{Example-UTG} and Appendix~\ref{GeneratedCases}. These examples show that our DMLM can help base models generate more reasonable, readable, fluent, and natural sentences.

\subsection{Computational Complexity}

Compared with vanilla RNN, our DM-RNN indeed increases the computational complexity from $O(T)$ to $O(T^2)$. In practice, we can follow \citet{PointerRNN} to set a context window that allows DMLM looks $L$ timesteps into the past at most, where $L$ is the context length. However, our DMLM can efficiently apply to Transformer-based models without additional computational complexity.

\begin{table*}[t]
	\begin{center}
	\resizebox{\textwidth}{!}{
		\begin{tabular}{l p{0.92\textwidth}}
			\toprule
			
	\textbf{Story context}: &  [FEMALE] bought packets of vegetable seeds from the store . she dug up the dirt in her garden . [FEMALE] planted onions , cilantro , and tomatoes . [FEMALE] watered the garden every night . \\\hline

\textbf{Golden Text:} & by the end of the summer [FEMALE] had enough vegetables to make salsa .\\\hline
	\hline
\textbf{PRPN}: & she got to work in the morning and was happy to have a garden . \\
\textbf{ON-LSTM}:  & [FEMALE] planted the plants and made it a huge success . \\ 
\textbf{AWD-LSTM}: &  [FEMALE] was happy to be helping her plants . \\
\textbf{DM-LSTM}: &  soon , [FEMALE] had enough vegetables to grow in her garden ! \\ \hline
\textbf{Transformer}:  &   she went to the store to buy the seeds . \\ 
\textbf{DM-Transformer}: &   soon , [FEMALE] had her garden full of vegetables ! \\ \hline
\textbf{GPT-2}: &   [FEMALE] 's garden grew very quickly and dry . \\

\textbf{DM-GPT-2}:  &  [FEMALE] now has fresh fruits and vegetables in her garden . \\ 
			\bottomrule
		\end{tabular}
	}
	\end{center}
		\caption{\label{Example-CTG}Examples of conditional text generation on ROCStories dataset.}
\end{table*}
\begin{table*}[t]
	\begin{center}
	\resizebox{\textwidth}{!}{
		\begin{tabular}{l p{0.92\textwidth}}
			\toprule
\textbf{Golden Text:} & what this group does is to take down various different websites it believes to be criminal and leading to terrorist acts . \\ \hline \hline
\textbf{PRPN}:	&
the right point to pay for the purchase of a bike , that ' s all we want to do to build , build together the support that i need to get here . 
\\ 
\textbf{ON-LSTM}:  & it ' s great to know that my experience has changed my mind because i ' m not going to work because i ' ve had to talk about that . \\

\textbf{AWD-LSTM}: & 
this is a tragic attack and it is understood that the pair will come up with a package of documents which may be possible . \\

\textbf{DM-LSTM}: & the win over bernie sanders was an emotional moment for clinton , who was running in the general election , though she lost their state of vermont .  \\
\hline
\textbf{Transformer}:   & ' i ' ve just been in that position so i ' ve never seen anything like this before , but it ' s something i have to say and i ' m going to go to and win this series . \\ 
\textbf{DM-Transformer}: &   in the second quarter of 2015 , the outlook for consumer spending rose 8 . 6 per cent , but for the fourth quarter , the company said it expects to expand by 0 . 7 per cent . \\\hline
\textbf{GPT-2}: &   if i had said a bit of pressure , i would probably be in a different position if i was a coach . \\ 
\textbf{DM-GPT-2}:  & they ' ve also said that it ' s difficult to know how many emails clinton actually sent to her in recent weeks or whether she would be the nominee . \\ 
			\bottomrule
		\end{tabular}
	}
	\end{center}
		\caption{\label{Example-UTG}Examples of unconditional text generation on EMNLP2017 WMT News dataset.}
\end{table*}

\section{Related Works}
Many previous studies have shown that leveraging the knowledge of syntactic structures can improve NLG~\cite{97Structured, 01Parsing, 05Syntactic, 15Dependency}. \citet{DependencyRNNforSentenceCompletion} incorporated syntactic dependencies into the RNN formulation, but they limited the scope to the scoring of complete sentences, not to next word prediction. Some other efforts have been done to integrate dependency structure into neural machine translation (NMT) from both the source and target side. \citet{Tree2SequenceNMT} proposed a tree-to-sequence attentional NMT model where source-side parse tree was used. \citet{seq2depNMT} involved target syntactic trees into NMT model to jointly learn target translation and dependency parsing. \citet{SIELM} introduced a syntactic inductive bias to NLG in an iterative non-autoregressive way.

For neural language models, recently, \citet{rnng} proposed recurrent neural network grammar (RNNG) to jointly model syntax and surface structure by incrementally generating a syntax tree and sentence. Subsequent work~\cite{urnng} extended the model to an unsupervised version. \citet{PRPN} introduced the Parsing-Reading-Predict Networks (PRPN) to calculate syntactic distances among words and use self-attention to compose previous states. Its subsequent work~\cite{onlstm} transferred the distance notion to LSTM cell, and introduced Ordered Neurons LSTM (ON-LSTM). 

However, all these methods, mainly based on RNN~\cite{rnn}, incorporate knowledge of syntactic structures by introducing complex architectural changes. Therefore, it can get very unwieldy to adapt them to other neural language models, such as Transformer and GPT-2.




\section{Conclusion}

In this paper, we introduce Dependency-based Mixture Language Models, which can incorporate knowledge of dependency structures into arbitrary auto-regressive generation models without any changes to the original architectures. Both automatic and human evaluation results in extensive experiments across different tasks and different architectures demonstrate the effectiveness and generalizability of our method.

In the future, we will explore to incorporate the dependency labels into our method, and combine our DMLM with more neural language models. Second, we would like to integrate other linguistic knowledge, such as constituency structures and semantic information, into neural language models in our manner.

\section*{Acknowledgements}
This work was supported by National Key R\&D Program of China (No.2018YFB1005100), Bejing Academy of Artificial Intelligence (BAAI) and State Key Laboratory of Media Convergence Production Technology and Systems. We appreciate the anonymous reviewers for their helpful comments. Xiaojun Wan is the corresponding author.

\bibliography{anthology,custom}

\begin{thebibliography}{55}
\expandafter\ifx\csname natexlab\endcsname\relax\def\natexlab#1{#1}\fi

\bibitem[{Ahmed et~al.(2019)Ahmed, Samee, and
  Mercer}]{YouOnlyNeedTraverseTrees}
Mahtab Ahmed, Muhammad~Rifayat Samee, and Robert~E. Mercer. 2019.
\newblock You only need attention to traverse trees.
\newblock In \emph{Proceedings of the 57th Conference of the Association for
  Computational Linguistics, {ACL} 2019, Florence, Italy, July 28- August 2,
  2019, Volume 1: Long Papers}, pages 316--322.

\bibitem[{Buys and Blunsom(2015)}]{15Dependency}
Jan Buys and Phil Blunsom. 2015.
\newblock Generative incremental dependency parsing with neural networks.
\newblock In \emph{Proceedings of the 53rd Annual Meeting of the Association
  for Computational Linguistics and the 7th International Joint Conference on
  Natural Language Processing of the Asian Federation of Natural Language
  Processing, {ACL} 2015, July 26-31, 2015, Beijing, China, Volume 2: Short
  Papers}, pages 863--869.

\bibitem[{Caccia et~al.(2020)Caccia, Caccia, Fedus, Larochelle, Pineau, and
  Charlin}]{GanFallingShort}
Massimo Caccia, Lucas Caccia, William Fedus, Hugo Larochelle, Joelle Pineau,
  and Laurent Charlin. 2020.
\newblock Language gans falling short.
\newblock In \emph{8th International Conference on Learning Representations,
  {ICLR} 2020, Addis Ababa, Ethiopia, April 26-30, 2020}.

\bibitem[{Casas et~al.(2020)Casas, Fonollosa, and Costa{-}juss{\`{a}}}]{SIELM}
Noe Casas, Jos{\'{e}} A.~R. Fonollosa, and Marta~R. Costa{-}juss{\`{a}}. 2020.
\newblock Syntax-driven iterative expansion language models for controllable
  text generation.
\newblock In \emph{Proceedings of the Fourth Workshop on Structured Prediction
  for NLP@EMNLP 2020, Online, November 20, 2020}, pages 1--10.

\bibitem[{Chelba(1997)}]{97Structured}
Ciprian Chelba. 1997.
\newblock A structured language model.
\newblock In \emph{35th Annual Meeting of the Association for Computational
  Linguistics and 8th Conference of the European Chapter of the Association for
  Computational Linguistics, Proceedings of the Conference, 7-12 July 1997,
  Universidad Nacional de Educaci{\'{o}}n a Distancia (UNED), Madrid, Spain},
  pages 498--500.

\bibitem[{Chung et~al.(2017)Chung, Ahn, and Bengio}]{HMRNN}
Junyoung Chung, Sungjin Ahn, and Yoshua Bengio. 2017.
\newblock Hierarchical multiscale recurrent neural networks.
\newblock In \emph{5th International Conference on Learning Representations,
  {ICLR} 2017, Toulon, France, April 24-26, 2017, Conference Track
  Proceedings}.

\bibitem[{Dai et~al.(2019)Dai, Yang, Yang, Carbonell, Le, and
  Salakhutdinov}]{transformer-xl}
Zihang Dai, Zhilin Yang, Yiming Yang, Jaime~G. Carbonell, Quoc~Viet Le, and
  Ruslan Salakhutdinov. 2019.
\newblock Transformer-xl: Attentive language models beyond a fixed-length
  context.
\newblock In \emph{Proceedings of the 57th Conference of the Association for
  Computational Linguistics, {ACL} 2019, Florence, Italy, July 28- August 2,
  2019, Volume 1: Long Papers}, pages 2978--2988.

\bibitem[{Devlin et~al.(2019)Devlin, Chang, Lee, and Toutanova}]{BERT}
Jacob Devlin, Ming{-}Wei Chang, Kenton Lee, and Kristina Toutanova. 2019.
\newblock {BERT:} pre-training of deep bidirectional transformers for language
  understanding.
\newblock In \emph{Proceedings of the 2019 Conference of the North American
  Chapter of the Association for Computational Linguistics: Human Language
  Technologies, {NAACL-HLT} 2019, Minneapolis, MN, USA, June 2-7, 2019, Volume
  1 (Long and Short Papers)}, pages 4171--4186.

\bibitem[{Du et~al.(2020)Du, Lin, Shen, O'Donnell, Bengio, and Zhang}]{DU-SYD}
Wenyu Du, Zhouhan Lin, Yikang Shen, Timothy~J. O'Donnell, Yoshua Bengio, and
  Yue Zhang. 2020.
\newblock Exploiting syntactic structure for better language modeling: {A}
  syntactic distance approach.
\newblock In \emph{Proceedings of the 58th Annual Meeting of the Association
  for Computational Linguistics, {ACL} 2020, Online, July 5-10, 2020}, pages
  6611--6628.

\bibitem[{Dyer et~al.(2016)Dyer, Kuncoro, Ballesteros, and Smith}]{rnng}
Chris Dyer, Adhiguna Kuncoro, Miguel Ballesteros, and Noah~A. Smith. 2016.
\newblock Recurrent neural network grammars.
\newblock In \emph{{NAACL} {HLT} 2016, The 2016 Conference of the North
  American Chapter of the Association for Computational Linguistics: Human
  Language Technologies, San Diego California, USA, June 12-17, 2016}, pages
  199--209.

\bibitem[{Emami and Jelinek(2005)}]{05Syntactic}
Ahmad Emami and Frederick Jelinek. 2005.
\newblock A neural syntactic language model.
\newblock \emph{Mach. Learn.}, 60(1-3):195--227.

\bibitem[{Eriguchi et~al.(2016)Eriguchi, Hashimoto, and
  Tsuruoka}]{Tree2SequenceNMT}
Akiko Eriguchi, Kazuma Hashimoto, and Yoshimasa Tsuruoka. 2016.
\newblock Tree-to-sequence attentional neural machine translation.
\newblock In \emph{Proceedings of the 54th Annual Meeting of the Association
  for Computational Linguistics, {ACL} 2016, August 7-12, 2016, Berlin,
  Germany, Volume 1: Long Papers}.

\bibitem[{Garg et~al.(2019)Garg, Peitz, Nallasamy, and
  Paulik}]{AlignTransformer}
Sarthak Garg, Stephan Peitz, Udhyakumar Nallasamy, and Matthias Paulik. 2019.
\newblock Jointly learning to align and translate with transformer models.
\newblock In \emph{Proceedings of the 2019 Conference on Empirical Methods in
  Natural Language Processing and the 9th International Joint Conference on
  Natural Language Processing, {EMNLP-IJCNLP} 2019, Hong Kong, China, November
  3-7, 2019}, pages 4452--4461.

\bibitem[{Guan et~al.(2020)Guan, Huang, Huang, Zhao, and Zhu}]{TACLStoryGen}
Jian Guan, Fei Huang, Minlie Huang, Zhihao Zhao, and Xiaoyan Zhu. 2020.
\newblock A knowledge-enhanced pretraining model for commonsense story
  generation.
\newblock \emph{Trans. Assoc. Comput. Linguistics}, 8:93--108.

\bibitem[{Guan and Huang(2020)}]{UNION}
Jian Guan and Minlie Huang. 2020.
\newblock {UNION:} an unreferenced metric for evaluating open-ended story
  generation.
\newblock In \emph{Proceedings of the 2020 Conference on Empirical Methods in
  Natural Language Processing, {EMNLP} 2020, Online, November 16-20, 2020},
  pages 9157--9166.

\bibitem[{Guo et~al.(2018)Guo, Lu, Cai, Zhang, Yu, and Wang}]{LeakyGAN}
Jiaxian Guo, Sidi Lu, Han Cai, Weinan Zhang, Yong Yu, and Jun Wang. 2018.
\newblock Long text generation via adversarial training with leaked
  information.
\newblock In \emph{Proceedings of the Thirty-Second {AAAI} Conference on
  Artificial Intelligence, (AAAI-18), the 30th innovative Applications of
  Artificial Intelligence (IAAI-18), and the 8th {AAAI} Symposium on
  Educational Advances in Artificial Intelligence (EAAI-18), New Orleans,
  Louisiana, USA, February 2-7, 2018}, pages 5141--5148.

\bibitem[{Hayes and Krippendorff(2007)}]{Krippendorff}
Andrew~F Hayes and Klaus Krippendorff. 2007.
\newblock Answering the call for a standard reliability measure for coding
  data.
\newblock \emph{Communication methods and measures}, 1(1):77--89.

\bibitem[{Hochreiter and Schmidhuber(1997)}]{LSTM}
Sepp Hochreiter and J{\"u}rgen Schmidhuber. 1997.
\newblock Long short-term memory.
\newblock \emph{Neural computation}, 9(8):1735--1780.

\bibitem[{Holtzman et~al.(2020)Holtzman, Buys, Du, Forbes, and Choi}]{topp}
Ari Holtzman, Jan Buys, Li~Du, Maxwell Forbes, and Yejin Choi. 2020.
\newblock The curious case of neural text degeneration.
\newblock In \emph{In 8th International Conference on Learning
  Representations}.

\bibitem[{Jacob et~al.(2018)Jacob, Lin, Sordoni, and
  Bengio}]{LearningHierarchicalStructures}
Athul~Paul Jacob, Zhouhan Lin, Alessandro Sordoni, and Yoshua Bengio. 2018.
\newblock Learning hierarchical structures on-the-fly with a
  recurrent-recursive model for sequences.
\newblock In \emph{Proceedings of The Third Workshop on Representation Learning
  for NLP, Rep4NLP@ACL 2018, Melbourne, Australia, July 20, 2018}, pages
  154--158.

\bibitem[{Kim et~al.(2019)Kim, Rush, Yu, Kuncoro, Dyer, and Melis}]{urnng}
Yoon Kim, Alexander~M. Rush, Lei Yu, Adhiguna Kuncoro, Chris Dyer, and
  G{\'{a}}bor Melis. 2019.
\newblock Unsupervised recurrent neural network grammars.
\newblock In \emph{Proceedings of the 2019 Conference of the North American
  Chapter of the Association for Computational Linguistics: Human Language
  Technologies, {NAACL-HLT} 2019, Minneapolis, MN, USA, June 2-7, 2019, Volume
  1 (Long and Short Papers)}, pages 1105--1117.

\bibitem[{Kong et~al.(2021)Kong, Huang, Tung, Guan, and Huang}]{StylizedStory}
Xiangzhe Kong, Jialiang Huang, Ziquan Tung, Jian Guan, and Minlie Huang. 2021.
\newblock Stylized story generation with style-guided planning.
\newblock In \emph{Findings of the Association for Computational Linguistics:
  {ACL/IJCNLP} 2021, Online Event, August 1-6, 2021}, volume {ACL/IJCNLP} 2021
  of \emph{Findings of {ACL}}, pages 2430--2436.

\bibitem[{Kuncoro et~al.(2018)Kuncoro, Dyer, Hale, Yogatama, Clark, and
  Blunsom}]{LSTMFailStructure}
Adhiguna Kuncoro, Chris Dyer, John Hale, Dani Yogatama, Stephen Clark, and Phil
  Blunsom. 2018.
\newblock Lstms can learn syntax-sensitive dependencies well, but modeling
  structure makes them better.
\newblock In \emph{Proceedings of the 56th Annual Meeting of the Association
  for Computational Linguistics, {ACL} 2018, Melbourne, Australia, July 15-20,
  2018, Volume 1: Long Papers}, pages 1426--1436.

\bibitem[{Li et~al.(2016)Li, Galley, Brockett, Gao, and Dolan}]{distinct}
Jiwei Li, Michel Galley, Chris Brockett, Jianfeng Gao, and Bill Dolan. 2016.
\newblock A diversity-promoting objective function for neural conversation
  models.
\newblock In \emph{{NAACL} {HLT} 2016, The 2016 Conference of the North
  American Chapter of the Association for Computational Linguistics: Human
  Language Technologies, San Diego California, USA, June 12-17, 2016}, pages
  110--119.

\bibitem[{Marcus et~al.(1993)Marcus, Santorini, and Marcinkiewicz}]{PTB1}
Mitchell~P. Marcus, Beatrice Santorini, and Mary~Ann Marcinkiewicz. 1993.
\newblock Building a large annotated corpus of english: The penn treebank.
\newblock \emph{Comput. Linguistics}, 19(2):313--330.

\bibitem[{Merity et~al.(2018)Merity, Keskar, and Socher}]{AWDLSTM}
Stephen Merity, Nitish~Shirish Keskar, and Richard Socher. 2018.
\newblock Regularizing and optimizing {LSTM} language models.
\newblock In \emph{6th International Conference on Learning Representations,
  {ICLR} 2018, Vancouver, BC, Canada, April 30 - May 3, 2018, Conference Track
  Proceedings}.

\bibitem[{Merity et~al.(2017)Merity, Xiong, Bradbury, and Socher}]{PointerRNN}
Stephen Merity, Caiming Xiong, James Bradbury, and Richard Socher. 2017.
\newblock Pointer sentinel mixture models.
\newblock In \emph{5th International Conference on Learning Representations,
  {ICLR} 2017, Toulon, France, April 24-26, 2017, Conference Track
  Proceedings}.

\bibitem[{Mikolov et~al.(2012)}]{PTB2}
Tom{\'a}{\v{s}} Mikolov et~al. 2012.
\newblock Statistical language models based on neural networks.
\newblock \emph{Presentation at Google, Mountain View, 2nd April}, 80:26.

\bibitem[{Mirowski and Vlachos(2015)}]{DependencyRNNforSentenceCompletion}
Piotr Mirowski and Andreas Vlachos. 2015.
\newblock Dependency recurrent neural language models for sentence completion.
\newblock In \emph{Proceedings of the 53rd Annual Meeting of the Association
  for Computational Linguistics and the 7th International Joint Conference on
  Natural Language Processing of the Asian Federation of Natural Language
  Processing, {ACL} 2015, July 26-31, 2015, Beijing, China, Volume 2: Short
  Papers}, pages 511--517.

\bibitem[{Mostafazadeh et~al.(2016)Mostafazadeh, Chambers, He, Parikh, Batra,
  Vanderwende, Kohli, and Allen}]{rocstories}
Nasrin Mostafazadeh, Nathanael Chambers, Xiaodong He, Devi Parikh, Dhruv Batra,
  Lucy Vanderwende, Pushmeet Kohli, and James~F. Allen. 2016.
\newblock A corpus and cloze evaluation for deeper understanding of commonsense
  stories.
\newblock In \emph{{NAACL} {HLT} 2016, The 2016 Conference of the North
  American Chapter of the Association for Computational Linguistics: Human
  Language Technologies, San Diego California, USA, June 12-17, 2016}, pages
  839--849.

\bibitem[{Papineni et~al.(2002)Papineni, Roukos, Ward, and Zhu}]{bleu}
Kishore Papineni, Salim Roukos, Todd Ward, and Wei{-}Jing Zhu. 2002.
\newblock Bleu: a method for automatic evaluation of machine translation.
\newblock In \emph{In Proceedings of the 40th Annual Meeting of the Association
  for Computational Linguistics}, pages 311--318.

\bibitem[{Peng et~al.(2019)Peng, Schwartz, and Smith}]{PaLM}
Hao Peng, Roy Schwartz, and Noah~A. Smith. 2019.
\newblock Palm: {A} hybrid parser and language model.
\newblock In \emph{Proceedings of the 2019 Conference on Empirical Methods in
  Natural Language Processing and the 9th International Joint Conference on
  Natural Language Processing, {EMNLP-IJCNLP} 2019, Hong Kong, China, November
  3-7, 2019}, pages 3642--3649.

\bibitem[{Radford et~al.(2019)Radford, Wu, Child, Luan, Amodei, Sutskever
  et~al.}]{gpt-2}
Alec Radford, Jeffrey Wu, Rewon Child, David Luan, Dario Amodei, Ilya
  Sutskever, et~al. 2019.
\newblock Language models are unsupervised multitask learners.
\newblock \emph{OpenAI blog}, 1(8):9.

\bibitem[{Roark(2001)}]{01Parsing}
Brian Roark. 2001.
\newblock Probabilistic top-down parsing and language modeling.
\newblock \emph{Comput. Linguistics}, 27(2):249--276.

\bibitem[{Semeniuta et~al.(2018)Semeniuta, Severyn, and Gelly}]{lm_rlm}
Stanislau Semeniuta, Aliaksei Severyn, and Sylvain Gelly. 2018.
\newblock On accurate evaluation of gans for language generation.
\newblock \emph{CoRR}, abs/1806.04936.

\bibitem[{Shen et~al.(2018)Shen, Lin, Huang, and Courville}]{PRPN}
Yikang Shen, Zhouhan Lin, Chin{-}Wei Huang, and Aaron~C. Courville. 2018.
\newblock Neural language modeling by jointly learning syntax and lexicon.
\newblock In \emph{6th International Conference on Learning Representations,
  {ICLR} 2018, Vancouver, BC, Canada, April 30 - May 3, 2018, Conference Track
  Proceedings}.

\bibitem[{Shen et~al.(2019)Shen, Tan, Sordoni, and Courville}]{onlstm}
Yikang Shen, Shawn Tan, Alessandro Sordoni, and Aaron~C. Courville. 2019.
\newblock Ordered neurons: Integrating tree structures into recurrent neural
  networks.
\newblock In \emph{7th International Conference on Learning Representations,
  {ICLR} 2019, New Orleans, LA, USA, May 6-9, 2019}.

\bibitem[{Shen et~al.(2021{\natexlab{a}})Shen, Tan, Sordoni, Reddy, and
  Courville}]{SOM}
Yikang Shen, Shawn Tan, Alessandro Sordoni, Siva Reddy, and Aaron~C. Courville.
  2021{\natexlab{a}}.
\newblock Explicitly modeling syntax in language models with incremental
  parsing and a dynamic oracle.
\newblock In \emph{Proceedings of the 2021 Conference of the North American
  Chapter of the Association for Computational Linguistics: Human Language
  Technologies, {NAACL-HLT} 2021, Online, June 6-11, 2021}, pages 1660--1672.

\bibitem[{Shen et~al.(2021{\natexlab{b}})Shen, Tay, Zheng, Bahri, Metzler, and
  Courville}]{StructFormer}
Yikang Shen, Yi~Tay, Che Zheng, Dara Bahri, Donald Metzler, and Aaron~C.
  Courville. 2021{\natexlab{b}}.
\newblock Structformer: Joint unsupervised induction of dependency and
  constituency structure from masked language modeling.
\newblock In \emph{Proceedings of the 59th Annual Meeting of the Association
  for Computational Linguistics and the 11th International Joint Conference on
  Natural Language Processing, {ACL/IJCNLP} 2021, (Volume 1: Long Papers),
  Virtual Event, August 1-6, 2021}, pages 7196--7209.

\bibitem[{Socher et~al.(2013)Socher, Perelygin, Wu, Chuang, Manning, Ng, and
  Potts}]{RNNforSentiment}
Richard Socher, Alex Perelygin, Jean Wu, Jason Chuang, Christopher~D. Manning,
  Andrew~Y. Ng, and Christopher Potts. 2013.
\newblock Recursive deep models for semantic compositionality over a sentiment
  treebank.
\newblock In \emph{Proceedings of the 2013 Conference on Empirical Methods in
  Natural Language Processing, {EMNLP} 2013, 18-21 October 2013, Grand Hyatt
  Seattle, Seattle, Washington, USA, {A} meeting of SIGDAT, a Special Interest
  Group of the {ACL}}, pages 1631--1642.

\bibitem[{Sutskever et~al.(2014)Sutskever, Vinyals, and Le}]{rnn}
Ilya Sutskever, Oriol Vinyals, and Quoc~V. Le. 2014.
\newblock Sequence to sequence learning with neural networks.
\newblock In \emph{Advances in Neural Information Processing Systems 27: Annual
  Conference on Neural Information Processing Systems 2014, December 8-13 2014,
  Montreal, Quebec, Canada}, pages 3104--3112.

\bibitem[{Takase et~al.(2018)Takase, Suzuki, and Nagata}]{DOC}
Sho Takase, Jun Suzuki, and Masaaki Nagata. 2018.
\newblock Direct output connection for a high-rank language model.
\newblock In \emph{Proceedings of the 2018 Conference on Empirical Methods in
  Natural Language Processing, Brussels, Belgium, October 31 - November 4,
  2018}, pages 4599--4609.

\bibitem[{Vaswani et~al.(2017)Vaswani, Shazeer, Parmar, Uszkoreit, Jones,
  Gomez, Kaiser, and Polosukhin}]{transformer}
Ashish Vaswani, Noam Shazeer, Niki Parmar, Jakob Uszkoreit, Llion Jones,
  Aidan~N. Gomez, Lukasz Kaiser, and Illia Polosukhin. 2017.
\newblock Attention is all you need.
\newblock In \emph{In Advances in Neural Information Processing Systems}, pages
  5998--6008.

\bibitem[{Wang et~al.(2019)Wang, Lee, and Chen}]{TreeTransformer}
Yau{-}Shian Wang, Hung{-}yi Lee, and Yun{-}Nung Chen. 2019.
\newblock Tree transformer: Integrating tree structures into self-attention.
\newblock In \emph{Proceedings of the 2019 Conference on Empirical Methods in
  Natural Language Processing and the 9th International Joint Conference on
  Natural Language Processing, {EMNLP-IJCNLP} 2019, Hong Kong, China, November
  3-7, 2019}, pages 1061--1070.

\bibitem[{Williams et~al.(2018)Williams, Drozdov, and Bowman}]{DoLatentTree}
Adina Williams, Andrew Drozdov, and Samuel~R. Bowman. 2018.
\newblock Do latent tree learning models identify meaningful structure in
  sentences?
\newblock \emph{Trans. Assoc. Comput. Linguistics}, 6:253--267.

\bibitem[{Wu et~al.(2017)Wu, Zhang, Yang, Li, and Zhou}]{seq2depNMT}
Shuangzhi Wu, Dongdong Zhang, Nan Yang, Mu~Li, and Ming Zhou. 2017.
\newblock Sequence-to-dependency neural machine translation.
\newblock In \emph{Proceedings of the 55th Annual Meeting of the Association
  for Computational Linguistics, {ACL} 2017, Vancouver, Canada, July 30 -
  August 4, Volume 1: Long Papers}, pages 698--707.

\bibitem[{Xu et~al.(2021)Xu, Guo, Tang, Su, Shou, Gong, Zhong, Quan, Jiang, and
  Duan}]{Syntax-EnhancedPTM}
Zenan Xu, Daya Guo, Duyu Tang, Qinliang Su, Linjun Shou, Ming Gong, Wanjun
  Zhong, Xiaojun Quan, Daxin Jiang, and Nan Duan. 2021.
\newblock Syntax-enhanced pre-trained model.
\newblock In \emph{Proceedings of the 59th Annual Meeting of the Association
  for Computational Linguistics and the 11th International Joint Conference on
  Natural Language Processing, {ACL/IJCNLP} 2021, (Volume 1: Long Papers),
  Virtual Event, August 1-6, 2021}, pages 5412--5422.

\bibitem[{Yang et~al.(2018)Yang, Dai, Salakhutdinov, and Cohen}]{MoS}
Zhilin Yang, Zihang Dai, Ruslan Salakhutdinov, and William~W. Cohen. 2018.
\newblock Breaking the softmax bottleneck: {A} high-rank {RNN} language model.
\newblock In \emph{6th International Conference on Learning Representations,
  {ICLR} 2018, Vancouver, BC, Canada, April 30 - May 3, 2018, Conference Track
  Proceedings}.

\bibitem[{Yu et~al.(2017)Yu, Zhang, Wang, and Yu}]{SeqGAN}
Lantao Yu, Weinan Zhang, Jun Wang, and Yong Yu. 2017.
\newblock Seqgan: Sequence generative adversarial nets with policy gradient.
\newblock In \emph{Proceedings of the Thirty-First {AAAI} Conference on
  Artificial Intelligence, February 4-9, 2017, San Francisco, California,
  {USA}}, pages 2852--2858.

\bibitem[{Zhang et~al.(2020)Zhang, Kishore, Wu, Weinberger, and
  Artzi}]{BERTScore}
Tianyi Zhang, Varsha Kishore, Felix Wu, Kilian~Q. Weinberger, and Yoav Artzi.
  2020.
\newblock Bertscore: Evaluating text generation with {BERT}.
\newblock In \emph{8th International Conference on Learning Representations,
  {ICLR} 2020, Addis Ababa, Ethiopia, April 26-30, 2020}.

\bibitem[{Zhang et~al.(2019)Zhang, Yang, Yuan, Shen, and Carin}]{SIVAE}
Xinyuan Zhang, Yi~Yang, Siyang Yuan, Dinghan Shen, and Lawrence Carin. 2019.
\newblock Syntax-infused variational autoencoder for text generation.
\newblock In \emph{Proceedings of the 57th Conference of the Association for
  Computational Linguistics, {ACL} 2019, Florence, Italy, July 28- August 2,
  2019, Volume 1: Long Papers}, pages 2069--2078.

\bibitem[{Zhou and Zhao(2019)}]{HPSGParser}
Junru Zhou and Hai Zhao. 2019.
\newblock Head-driven phrase structure grammar parsing on penn treebank.
\newblock In \emph{Proceedings of the 57th Conference of the Association for
  Computational Linguistics, {ACL} 2019, Florence, Italy, July 28- August 2,
  2019, Volume 1: Long Papers}, pages 2396--2408.

\bibitem[{Zhu et~al.(2018)Zhu, Lu, Zheng, Guo, Zhang, Wang, and Yu}]{self-bleu}
Yaoming Zhu, Sidi Lu, Lei Zheng, Jiaxian Guo, Weinan Zhang, Jun Wang, and Yong
  Yu. 2018.
\newblock Texygen: {A} benchmarking platform for text generation models.
\newblock In \emph{The 41st International {ACM} {SIGIR} Conference on Research
  {\&} Development in Information Retrieval}, pages 1097--1100.

\bibitem[{Zilly et~al.(2017)Zilly, Srivastava, Koutn{\'{\i}}k, and
  Schmidhuber}]{RHN}
Julian~Georg Zilly, Rupesh~Kumar Srivastava, Jan Koutn{\'{\i}}k, and
  J{\"{u}}rgen Schmidhuber. 2017.
\newblock Recurrent highway networks.
\newblock In \emph{Proceedings of the 34th International Conference on Machine
  Learning, {ICML} 2017, Sydney, NSW, Australia, 6-11 August 2017}, volume~70,
  pages 4189--4198.

\bibitem[{Zolna et~al.(2018)Zolna, Arpit, Suhubdy, and
  Bengio}]{FraternalDropout}
Konrad Zolna, Devansh Arpit, Dendi Suhubdy, and Yoshua Bengio. 2018.
\newblock Fraternal dropout.
\newblock In \emph{6th International Conference on Learning Representations,
  {ICLR} 2018, Vancouver, BC, Canada, April 30 - May 3, 2018, Conference Track
  Proceedings}.

\end{thebibliography}
\bibliographystyle{acl_natbib}

\clearpage
\appendix

\section{Experimental Setup}
\label{DetailSetup}
All the algorithms are implemented in Pytorch and trained on a machine with 8 NVIDIA GTX 2080Ti GPUs.

\subsection{Conditional Text Generation}
The dataset statistics of ROCStories dataset is reported in Table~\ref{Stat-ROC}.
\begin{table}[h]
\begin{center}
\begin{tabular}{c|ccc}
\toprule
           & Train       & Validation   & Test    \\ \hline
\#Stories & 78,529      & 9,816      & 9,816      \\
\bottomrule
\end{tabular}
\end{center}
\caption{Statistics of ROCStories dataset.}
\label{Stat-ROC}
\end{table}

In this task, both the DM-LSTM and base LSTM are built on a AWD-LSTM language model with an embedding size of $400$ and hidden layer units $1150$. The dropout rates are $0.4, 0.25, 0.4$ for the output of the last layer, outputs between
LSTM layers, and input embedding layers, respectively. The weight dropout for the RNN hidden to hidden matrix is $0.5$, and the dropout rate to remove words from embedding layer is $0.1$. The context length for DM-LSTM is set to $56$. For PRPN and ON-LSTM, we keep their original settings. 

In this task, all the models are trained on a singe GPU with learning rate $30$, weight decay $1.2e-6$. LSTM baselines are trained for $500$ epochs with batch size $100$. DM-LSTM is first trained by dependency modeling objective for $100$ epochs with batch size $80$, and then by language modeling in Equation~\ref{eq-MLE-finetune} for $400$ epochs with batch size $60$ due the computational budgets limit.
 
 For both the DM-Transformer and base Transformer, we use a standard $6$-layer Transformer language model with $8$ attention heads, embedding dimension $512$, projection dimension $2048$ and dropout rate $0.1$. During training, we use Adam optimizer with $\beta_{1} = 0.9$, $\beta_{2} = 0.98$, weight decay $0.01$ and learning rate $5e-4$, and apply the dynamic batching provided by fairseq\footnote{\url{https://github.com/pytorch/fairseq}} to train both the models with 4 GPUs. Transformer is trained for $60$ epochs, while DM-GPT-2 is first trained by dependency modeling for $30$ epochs, and then trained by language modeling in Equation~\ref{eq-MLE-finetune} for $30$ epochs.

We use the pretrained GPT-2-base model for both the DM-GPT-2 and base GPT-2. In this comparison, we apply the same training settings with Transformer-base models except that learning rate is set to $5e-5$. GPT-2 is trained for $80$ epochs, while DM-GPT-2 is first trained by dependency modeling for $40$ epochs, and then trained by language modeling in Equation~\ref{eq-MLE-finetune} for $40$ epochs.

For all the models, we select the best checkpoint according to the loss of validation set for testing.

\subsection{Unconditional Text Generation}
The dataset statistics of EMNLP2017 WMT News dataset is reported in Table~\ref{Stat-News}.
\begin{table}[h]
\begin{center}
\begin{tabular}{c|ccc}
\toprule
           & Train       & Validation   & Test    \\ \hline
\#Stories & 268,586      & 10,000      & 10,000      \\
\bottomrule
\end{tabular}
\end{center}
\caption{Statistics of EMNLP2017 WMT News dataset.}
\label{Stat-News}
\end{table}

The context length for DM-LSTM is set to $36$. LSTM baselines are trained for $500$ epochs with batch size $300$. DM-LSTM is first trained by dependency modeling objective for $100$ epochs with batch size $300$, and then by language modeling for $400$ epochs with batch size $200$. Besides, all the other experimental setups are the same with those for the conditional text generation task.  

\subsection{Language Modeling}
The dataset statistics of Penn Treebank dataset is reported in Table~\ref{Stat-PTB}.
\begin{table}[h]
\begin{center}
\begin{tabular}{c|ccc}
\toprule
           & Train       & Validation   & Test    \\ \hline
\#Stories & 42,068      & 3,370      &  3,761     \\
\bottomrule
\end{tabular}
\end{center}
\caption{Statistics of Penn Treebank dataset.}
\label{Stat-PTB}
\end{table}

The context length for DM-LSTM is set to $16$. DM-LSTM is trained for 1000 epochs with batch size $20$, following~\cite{AWDLSTM}. Besides, all the other experimental setups are the same with those for the conditional text generation task.

\section{Impact of the Dependency Parser}
\label{impact_parser}
In our work, we use an off-the-shelf dependency parser to get the dependency parse trees for dependency modeling. Consequently, the better the quality of dependency parsing, the better the performance of our method. HPSG Parser~\cite{HPSGParser}, the dependency parser we use, is one of the state-state-of-the-art parsers. This ensures the high quality of parsing results. \citet{HPSGParser} trained HPSG Parser with the training set of PTB, and kept the test set held-out. So, when we do language modeling on PTB, the parser will not inject any future predictions that contribute to testing. 

HPSG Parser maintains high-quality on out-of-domain text, as shown in its paper~\cite{HPSGParser}. Most importantly, even on the out-of-domain datasets, i.e., ROCStories and EMNLP2017 WMT News, our work can still obtain a significant improvement, as shown in Section~\ref{CTG} and Section~\ref{UTG}.

\section{Language Modeling on Transformer-based Models}
\label{Transformer-lm}
The language modeling results of Transformer-based models evaluated on PTB dataset are shown in following Table~\ref{tab:lm-transformer-ppl}.
\begin{table}[h]
\centering
\resizebox{\columnwidth}{!}{
\begin{tabular}{l|lll}
\toprule
Models                & \#Params & Dev PPL & Test PPL \\ \hline
Transformer & 24M      & 100.7   & 106.7     \\
DM-Transformer & 24M      & \textbf{80.6}   &  \textbf{84.6}     \\ \hline
GPT-2 & 163M & 62.6 & 55.2 \\
DM-GPT-2 & 163M & \textbf{58.8} & \textbf{51.6} \\
\bottomrule
\end{tabular} }
\caption{\label{tab:lm-transformer-ppl}  Transformer-based models' perplexity evaluated on validation and test sets of Penn Treebank dataset.}
\end{table}

The good performance of Transformer-based models often rely on training with large datasets, but PTB is a very small dataset. Therefore, Transformer-based models perform worse than LSTM-based models, as shown in Table~\ref{tab:lm-ppl} and Table~\ref{tab:lm-transformer-ppl}. However, our DM-models still substantially reduce the perplexity compared with base models. DM-Transformer improves the base Transformer by over $20$ perplexity points on both the validation and test set, and DM-GPT-2 also improves the base GPT-2 by almost $4$ perplexity points. These results further confirm the effectiveness our method. 

\section{Human Evaluation}
\label{HEQuestion}

We post the human evaluation questionnaire, as shown in Table~\ref{Question-CTG} and Table~\ref{Question-UTG}, and then recruit five workers with sufficient high English skills. We pay each worker 45 US dollars, and let them complete the evaluation within a week.

\section{Generated Examples}
\label{GeneratedCases}
For a more general comparison, we present more generated examples of unconditional text generation in Table~\ref{MoreExample-UTG}.

\begin{table*}[t]
	\begin{center}
	\resizebox{\textwidth}{!}{
		\begin{tabular}{p{\textwidth}}
			\toprule
			
			\textbf{Task Description}

Each story contains about five sentences. For each story, we will put the first four sentences into two different systems, and then systems generate the last sentence. The requirement for this manual evaluation is to judge \textbf{which story better complies with the English grammar norm, and is more logically related to the first four sentences.} 

\textbf{NOTE} that the names in all stories are replaced with "[MALE]" or "[FEMALE]" or "[NEUTRAL]", and all the sentences are preprocessed by lowercasing, separating punctuation, and splitting conjunctions. They are not grammar errors. Please ignore these when evaluating and do not allow them to affect your judgments.
\\ \hline
\textbf{Evaluation Criterion}

You need to compare the stories from two metrics: \textbf{grammaticality} and \textbf{logicality}. And the two metrics are \textbf{independent} of each other. One of the judgments should not have any influence on the other one. Specific criteria for evaluating are as follows:

\textbf{1. Grammaticality}

In the process of evaluating grammaticality, it should be considered whether the statement itself complies with the English standard usage. Then annotate which story is better at grammaticality. You may not care about what the generated sentences are saying but \textbf{only if there are any grammatical problems in the sentence itself.}

\textbf{2. Logicality}

In the process of evaluating logicality, you need to carefully read the whole story including the first four sentences and the generated sentence, and compare stories in logicality. Then annotate which story is better at logicality in terms of the coherence to the given beginnings and the inter-sentence causal and temporal dependencies. In this process, you may encounter sentences that are not completely grammatical.\textbf{Please make a logical evaluation based on the main part of the sentence (such as some keywords, etc.) and what you can intuitively feel.} Under the circumstances, the story can be judged totally illogical only if the grammar is too poor to understand the meaning or the logic is unreasonable.
\\  \hline
\textbf{Notes}

· Again, the grammaticality and logicality of the story are \textbf{two independent metrics}. Some very logically inappropriate generated stories are good in the grammaticality part, and there are some stories with obvious grammatical errors but they don't affect the respective judgment.

· Sometimes, there may be more than one kind of reasonable story for a beginning. Please do not limit your imagination. \textbf{As long as the story is logically reasonable, direct, and able to make sense, it can be judged good in logicality.}

· Some stories may not be accurately judged. In the process of determining the comparison of this type of two stories, according to your own understanding of the examples and the subjective feelings of the stories, choose a better story you think is the most appropriate. \textbf{Please ensure that your evaluation criterion for different stories is the same.} \\
			\bottomrule
		\end{tabular}
	}
	\end{center}
		\caption{\label{Question-CTG}Human evaluation questionnaire for conditional text generation.}
\end{table*}

\begin{table*}[t]
	\begin{center}
	\resizebox{\textwidth}{!}{
		\begin{tabular}{p{\textwidth}}
			\toprule
			
	\textbf{Task Description}\\

In this review, you will read 900 sentences. For each sentence, you should determine \textbf{whether the sentence is written by human}.
\textbf{Note}: All the sentences are preprocessed by lowercasing, separating punctuation, and splitting conjunctions. They are not grammar errors. Some sentences may have a specific context, or they may be talking about completely fictitious things. Please ignore these when evaluating and do not allow them to affect your judgments.

\textbf{Evaluation Criterion}

The judgment can mainly depend on your own understanding and the subjective feelings. But fluency, readability, engagement (whether you felt interested about the sentence), and anything else that you think is important can also help you make a decision. \\
			\bottomrule
		\end{tabular}
	}
	\end{center}
		\caption{\label{Question-UTG}Human evaluation questionnaire for unconditional text generation.}
\end{table*}

\begin{table*}[t]
	\begin{center}
	\resizebox{\textwidth}{!}{
		\begin{tabular}{l p{0.97\textwidth}}
			\toprule
\multirow{6}{*}{\textbf{Golden Text:}}
 & over 1 , 600 a day have reached greece this month , a higher rate than last july when the crisis was already in full swing .\\\cline{2-2} 
 & " we ' re working through a legacy period , with legacy products that are 10 or 20 years old , " he says .\\\cline{2-2} 
 & ' the first time anyone says you need help , i ' m on the defensive , but that ' s all that i know .\\\cline{2-2} 
 & out of those who came last year , 69 per cent were men , 18 per cent were children and just 13 per cent were women .\\\hline
	\hline
\multirow{7}{*}{\textbf{PRPN: }}
& as a mother , i can ' t work to be working on some kind of stuff , but i ' m not really sure that the single market is going to be as bad as i ' m on . \\\cline{2-2} 
& in fact , there is a good position to focus on this and that will be a clear opportunity for the us to make sure that we do not have any concerns . \\\cline{2-2} 
& there ' s still more opportunities than that , but this is what you ' re talking about , but it ' s not right . \\\cline{2-2}  
& as well as a labour party , the former party member who claimed the vote in the referendum on whether to vote to leave the eu should be questioned . \\ \hline
\multirow{7}{*}{\textbf{ON-LSTM: }}
 & so they did that because we ' ve been saying they ' re going to be fighting for this state , but they ' re going to keep going . \\\cline{2-2} 
 & the official said they were hoping to make a contribution in its strong inflation growth in the future , and that a more conservative leader could look for jobs and be stronger . \\\cline{2-2} 
 & it ' s something that i think are a good team , the first place to do it and i ' m really happy . \\\cline{2-2} 
 &  ' there ' s no question that the person we ' re going to take is probably an important thing to be asked , " said john . \\
\hline
\multirow{5}{*}{\textbf{AWD-LSTM: }}
& in this month ' s election , the u . s . economy has fallen in the past few years , a higher than a decade ago . \\\cline{2-2}
& in the last year i had been an 18 - year - old woman in my two - year - old son . \\\cline{2-2}
& it was a great test for me to try to get back on the bench and be there , it ' s a huge challenge for us . \\\cline{2-2}
& i just think it ' s important for us to do something that would help them in the best way we can to do it . 
\\
\hline
\multirow{6}{*}{\textbf{DM-LSTM: }}
 & " the united states has to come to mind that the threat of climate change is less of a serious issue , " the pentagon said in a statement . \\\cline{2-2} 
 & in the event of an initial campaign for the democratic nomination , he had released some of the most controversial ads that they had been speaking about since he was a president .  \\\cline{2-2} 
 & there is an example of a presidential candidate who has been on the debate trail for more than a year .  \\\cline{2-2} 
 & the central bank of japan is set to raise its benchmark interest rate at its first time in nearly a decade . \\\hline
\hline
\multirow{7}{*}{\textbf{Transformer}:}
 & you can ' t get away with things that are better than you did at home and hopefully get better than not the first team . \\\cline{2-2} 
& in the case of the cases , the nsw government said it would accept 10 , 000 additional emergency costs if it did not help the industry . \\ \cline{2-2} 
& if there is an oil price that is at stake , it is not as far as the price of oil .\\ \cline{2-2} 
& the country has promised to build a nationwide population of about 150 , 000 to more than 2 , 000 , with a budget to help in building more affordable housing .\\ \hline
\multirow{7}{*}{\textbf{DM-Transformer}:}
 & in this particular area , as in the modern world , he is seen as someone who takes the risk of suffering a heart attack . \\ \cline{2-2} 
& that ' s why we ' re talking about the second half of the year , and a lot of people have asked us to do the best we can . \\ \cline{2-2} 
& the vast majority of american voters , particularly those who chose trump , said that he had changed the result . \\ \cline{2-2} 
& so this is a big step , and i ' m really excited to be part of the new york olympics . \\ \hline \hline
\multirow{8}{*}{\textbf{GPT-2}:}
& the reason is that the student community who doesn ' t know what he ' s talking about , or who ' s not even a businessman , he ' s going to take care of itself . \\ \cline{2-2}
&  the difference is that the reality of " brexit " has been the single largest trading partner in the world , and now is it . \\ \cline{2-2} 
& the game is now used to push for players to learn from them and learn from them and also play in the front of them . \\ \cline{2-2} 
&  the first woman to run for president is to make a case for a woman she wants to make as president of the united states . 
\\ \hline
\multirow{6}{*}{\textbf{DM-GPT-2}:}
& " i just thought that the whole picture was a strange story , " he said in a telephone interview on thursday . \\ \cline{2-2} 
&  " the importance of local authorities is very strong , " she said in an interview on friday afternoon .\\ \cline{2-2} 
&  we are working closely with the government to resolve this issue and have to work with local authorities to resolve the problem .\\ \cline{2-2} 
&  a final verdict will be held on thursday at the supreme court in washington on march 15 , 2017 . \\ 
			\bottomrule
		\end{tabular}
	}
	\end{center}
		\caption{\label{MoreExample-UTG}Examples of unconditional text generation on EMNLP2017 WMT News dataset.}
\end{table*}

\end{document}